\documentclass{ecai}
\usepackage{times}
\usepackage{graphicx}
\usepackage{multirow}
\usepackage{latexsym}
\usepackage{xcolor}
\usepackage{xspace}
\usepackage[linesnumbered,ruled,vlined]{algorithm2e}
\usepackage{array}
\usepackage{xcolor}
\usepackage{xspace}
\usepackage{comment}
\usepackage{float}
\restylefloat{table}

\newcommand{\MethodName}{RankML\xspace}

\begin{document}

\title{RankML: Meta Learning-Based Approach for Pre-Ranking Machine Learning Pipelines}

\author{Doron Laadan \and Roman Vainshtein \and Yarden Curiel \and Gilad Katz \and Lior Rokach\institute{Ben-Gurion University of the Negev, \{laadan,romanva,curiey\}@post.bgu.ac.il,\{giladkz,liorrk\}@bgu.ac.il} }

\maketitle
\bibliographystyle{ecai}

\begin{abstract}
 The explosion of digital data has created multiple opportunities for organizations and individuals to leverage machine learning (ML) to transform the way they operate. However, the shortage of experts in the field of machine learning -- data scientists -- is often a setback to the use of ML. In an attempt to alleviate this shortage, multiple approaches for the automation of machine learning have been proposed in recent years. While these approaches are effective, they often require a great deal of time and computing resources. In this study, we propose \MethodName, a meta-learning based approach for predicting the performance of whole machine learning pipelines. Given a previously-unseen dataset, a performance metric, and a set of candidate pipelines, \MethodName immediately produces a ranked list of all pipelines based on their predicted performance. Extensive evaluation on 244 datasets, both in regression and classification tasks, shows that our approach either outperforms or is comparable to state-of-the-art, computationally heavy approaches while requiring a fraction of the time and computational cost.
\end{abstract}
\section{Introduction}
Machine learning (ML) has been successfully used in a broad range of applications, including recommender systems \cite{covington2016deep}, anomaly detection \cite{chandola2009anomaly}, and social networks analysis \cite{wang2015link}. This trend has been driven by the enormous growth in the creation of digital data, which enables organizations to analyze and derive insights from almost every aspect of their activities. The growth in the use of ML, however, has not been accompanied by a similar growth in the number of human experts capable of applying it, namely data scientists \cite{davenport2012data}.

To overcome the shortage in skilled individuals, multiple approaches for automatic machine learning (AutoML) have been proposed in recent years. While earlier studies focused on specific tasks in the ML pipeline -- hyperparameter tuning \cite{bergstra2012random}, feature engineering and feature selection \cite{khalid2014survey}, etc. -- recent studies such as \cite{Drori2018b,Olson2019b,Thornton2012b,Feurer2015d} seek to automate the creation of the entire ML pipeline end-to-end.

Despite its large diversity, both in the modeling of the problem and in the algorithms used, the field of automatic ML pipeline generation is computationally expensive as well as time-consuming. The reasons for these shortcomings include a very large search space both for algorithms and pipeline architectures, the need to perform hyper-parameter optimization, and the fact that evaluating even a single pipeline on a very large dataset may require hours. Another significant shortcoming of most existing approaches is their inability to learn from previously analyzed datasets, which forces them to start ``from scratch'' with every new dataset. Few approaches such as \cite{Feurer2015d,fusi2018probabilistic} do try to utilize previous knowledge but do so on a limited scope and mostly during initialization.

In this study, we present \MethodName, a novel meta learning-based approach for ML pipeline performance prediction. Given a dataset, a set of candidate pipelines, an evaluation task (e.g., classification, regression) and an evaluation metric, \MethodName produces a ranked list of all candidate pipelines based on their expected performance with regard to the metric. This list is produced solely based on meta-knowledge gained from previously-analyzed datasets and pipelines combinations, without running any of the candidate pipelines. As a result, \MethodName has a computational complexity near O(1). 

We compare the performance of \MethodName to those of current state-of-the-art pipeline generation approaches---the TPOT \cite{Olson2019b} and auto-sklearn\cite{Feurer2015d} frameworks. The results of the evaluation, conducted both on classification and regression problems, show that our approach achieves better or comparable (BOC) results to the baselines at a fraction of the time.

Our contributions in this paper are as follows:
\begin{itemize}
    \item We present \MethodName, a meta learning-based approach for the ranking of ML pipeline based on their predicted performance. \MethodName leverages insights from previously analyzed datasets and pipelines combinations and is therefore capable of producing predictions without running on the current datasets.
    \item We propose a novel meta-learning approach for pipeline analysis. We derive meta-features both from the analyzed dataset and the pipeline's topology, and demonstrate that this combination yields state-of-the-art results.
    \item We generate a large, publicly available dataset of pipelines and their performance results on multiple datasets. The information is available both for classification and regression tasks.
\end{itemize}

\section{Related Work}
\subsection{Automated machine learning}
Automated machine learning (AutoML) is the process of automating the application of machine learning to real-world problems, without human intervention. The goal of this field of research is usually to enable non-experts to effectively utilize "off the shelf" solutions or save time and effort for knowledgeable practitioners.

At its core, the problem  AutoML is trying to solve is as follows: given a dataset, a machine learning task and a performance criterion, solve the task with respect to the dataset while optimizing the performance \cite{Drori2018b}. Finding an optimal solution is especially challenging due to the growing amount of machine learning models available and their hyper-parameters configurations, which can severely affect the performances of the model \cite{Luo2016b,Thornton2012b}.

Multiple approaches have been proposed to tackle the above problem. These approaches range from automatic feature engineering \cite{Katz2017b} to automatic model selection \cite{Vainshtein2018}. Some approaches attempt to automatically and simultaneously choose a learning algorithm and optimize its hyper-parameters. This approach is also known as \textit{combined algorithm
selection and hyperparameter optimization problem} (CASH) \cite{Thornton2012b,Feurer2015e}. More recently, several studies \cite{Drori2018b,Olson2019b} proposed the automation of the entire work-flow, building a complete machine learning pipeline for a given dataset and task.

Automating the creation of entire ML pipelines is difficult due to the extremely large search space, both of the pipeline architecture and the algorithms that populate it. Furthermore, the fact that the performance of each algorithm is highly dependent on the input it receives from the previous algorithm(s) adds another dimension of complexity. To overcome this challenge, different studies propose a large range of approaches \cite{milutinovic2017end}. TPOT and Autostacker  \cite{Olson2019b,chen2018autostacker} for example, use genetic programming to create and evolve the pipelines while auto-weka and auto-sklearn \cite{Thornton2012b,Feurer2015e} use Bayesian Optimization to solve the CASH problem. Another recent approach is used by autoDi \cite{Vainshtein2018}, which applies word embedding of domain knowledge gathered from academic publications and dataset meta-features to recommend a suitable algorithm.

In the majority of cases, most of the above-mentioned pipeline creation methods perform well and produce high, competitive performance results. However, most studies in the field suffer from two main shortcomings. First and foremost, applying these approaches is very computationally expensive, with running times that can easily reach days for large datasets \cite{Luo2016b,Olson2019b,Thornton2012b}. The second shortcoming is that most state-of-the-art methods are not sufficiently generic and rely on their underlying code packages to run (e.g., the use of scikit-learn for auto-sklearn and TPOT). This limitation may prevent automatic pipeline generation frameworks from integrating unique types of primitives not implemented in their respective underlying code packages.

Noticeably, several studies propose (albeit partial) solutions to these two challenges. AlpahD3M \cite{Drori2018b} strives to use a broad set of primitives to synthesize a pipeline and set the appropriate hyper-parameters regardless of the underlying code packages. AutoDi and autoGRD \cite{Noy2019} generates an offline model (i.e., one that is fully-trained prior to its application to new datasets) that can be applied almost instantly at runtime. In addition, auto-sklearn uses a meta-learning approach to decrease the time of the Bayesian optimization problem \cite{Feurer2015d} and most recently the use of portfolio-based algorithms to improve the process runtime\cite{feurer2018practical}. Additional solutions are also proposed in the literature such as \cite{fusi2018probabilistic} that uses a collaborative filtering approach or \cite{jin2019auto} which tackle the issue of automating the neural architecture search (NAS) problem.

\subsection{Meta-learning}
Meta-learning, or learning to learn, is commonly used to describe the scientific approach of observing different machine learning algorithms performances on a range of learning tasks. We then use those observations -- the meta-data -- to learn a new task or to improve an existing algorithm's performance. Simply put, meta-learning is the process of understanding and adapting learning itself on a higher level \cite{Lemke2015b}. Instead of starting `from scratch', we leverage previously-gained insights.

Throughout the years, multiple studies have been exploring the application of meta-learning in various domains such as ensemble methods  \cite{brazdil2008metalearning,wolpert2002supervised,doi:10.1002/widm.1249}, algorithms recommendations \cite{Vainshtein2018,brazdil2003ranking}, meta-learning systems and transfer learning \cite{Vilalta2002b}. Many state-of-the-art AutoML methods use meta-learning as a way of improving their accuracy and speed \cite{Feurer2015d,Santoro2016b,feurer2018scalable} and multiple studies describe ways to create meta knowledge usable by machine learning algorithms \cite{brazdil2003ranking,Katz2017b}.

Meta knowledge usually involves creating significant and meaningful meta-features on the datasets or the models used  \cite{Feurer2015d,Katz2017b,Vainshtein2018,Noy2019}. The majority of meta-features can be divided into five ``families'', as shown in \cite{Feurer2015d}. Examples of such families include Landmarking \cite{Pfahringer2000b}, which achieves state-of-the-art results but is computationally-heavy. Another example is the derivation of meta-features on performance, which is easy to extract but does not necessarily yield optimal results \cite{Lemke2015b}. The design of such meta-features, although an important process, is also considered a challenge for meta-learning \cite{brazdil2008metalearning,brazdil2003ranking}. For that reason, In recent years several frameworks were proposed for the automatic extraction of meta-features such as as\cite{10.1007/978-3-319-31753-3_18}.

Recently, several studies proposed the use of meta-features and learning to improve the AutoML process. AutoDi \cite{Vainshtein2018} and AutoGRD  \cite{Noy2019} , for example, use meta-features of datasets to rank different machine learning algorithms and both already achieved good results in model recommendation task. Katz et al. \cite{Katz2017b}, used meta-features for automatic feature engineering. Our approach will utilize the dataset meta-features with a combination of pipelines topology. It will be, to the best of our knowledge, the first time such combination is explored.

\section{Problem Formulation}
We use the same notations and definitions as previous works in this field, specifically \cite{Olson2019b} and \cite{Drori2018b}:\newline
\noindent \textbf{Primitives.} A set of different algorithms that can be applied to a dataset as part of the data mining work-flow. We divide primitives into four families:
\begin{itemize}
    \item \textbf{Data pre-processing.} Consisting of algorithms for data cleaning, balancing, resampling, label encoding, and missing values imputation.
    \item \textbf{Feature pre-processing.} Consisting of algorithms such as Principal Components Analysis (PCA) and Synthetic Minority Over-sampling Technique (SMOTE).
    \item \textbf{Features engineering and extraction.} Consists of algorithms used for discretization and feature engineering.
    \item \textbf{Predictive models.} Consists of all algorithms used to produce a final prediction. This family includes algorithms for classification, regression, ranking, etc. Relevant deep architectures are also included in this family.
\end{itemize}
\vspace{6pt}

\noindent \textbf{Pipeline.} A directed acyclic graph (DAG) $G=\{V,E\}$, where the vertices of the graph are \textit{primitives} and the edges of the graph determine the primitives' order of activation and input. The pipeline constitutes a complete data mining work-flow: its design can be a simple one-way DAG in which each primitive output is the input to the next primitive or it could have a more complex design in which the input for a primitive is the concatenation of the output of several primitives. \newline 

\noindent \textbf{Objective function.} We define an \textit{AutoML job} $\mathcal{T}(D,T,M)$ consisting of tabular dataset $D$ with $m$ columns and $k$ instances, a machine learning task $T$ and a performance metric $M$. Additionally, we assume a list of candidate machine learning pipelines, or \textit{pipelines} $\mathcal{C}=\{c_1,c_2 \dots, c_{N_c}\}$ for the given AutoML job. Given that pipeline $c \in \mathcal{C}$ is able to produce predictions over the specific AutoML job $\mathcal{T}$, our goal is to produce an ordered list of the candidates pipelines $\mathcal{C}_{Ranked}$, order by the following function: 
\[\arg\min_{c \in \mathcal{C}} \mathcal{E}(\mathcal{T}(D, T, M), c)\]
Where $\mathcal{E}$ is the error of pipeline $c$ over the AutoML job $\mathcal{T}$.
\section{The Proposed Method}
\noindent \textbf{Overview.} Our process is presented in Figure \ref{fig:MetaLearning_workflow}. It consists of an \textit{offline} and an \textit{online} phase. In the offline phase, we generate and train multiple pipeline architectures on a large set of diverse datasets and record their performance. Also, we extract meta-features that model the dataset, the pipeline, and their interdependence. We then use these meta-features to train a ranking algorithm capable of predicting the final performance of given dataset-pipeline combinations without actually running them.

In the online phase, \MethodName receives a previously unseen dataset, a set of candidate pipelines and an evaluation metric. We then extract meta-features describing both the dataset and each of the candidate pipelines and use the ranking algorithm to produce a ranked list of the candidate pipelines. Next, the top-ranked pipelines are evaluated. Finally, the actual performance is recorded and added to our knowledge-base for future use.
  
It is important to point out that \MethodName is not limited in any way in its sources for candidate pipelines. More specifically, the pipelines ranked by \MethodName can be randomly generated or received from other pipeline generation frameworks. In this sense, our approach can function both as a stand-alone ML pipeline recommendation framework and as a preliminary step for other, more computationally intensive solutions.

In the remainder of this section we present the processes we use to extract the various meta-features used by our model. We then describe the process of training the meta-model.

\subsection{Dataset Meta-Features}
To create our dataset meta-features we build upon the previous work of \cite{Katz2017b} and \cite{Vainshtein2018}, who successfully used dataset-based meta-features for AutoML related tasks. Our meta-features combine elements from both studies and can be divided into two groups:
\begin{itemize}
    \item \textbf{Descriptive.} Used to describe various aspects of the dataset. This group of meta-features includes information such as the number of instances in the data, number of attributes, percentage of missing values and likewise.
    \item \textbf{Correlation-based.} Used to model the interdependence of features within the analyzed datasets. Meta-features of this group includes the correlation between different attributes and the target value, Pearson correlation between attributes and different aggregations such as average and standard deviation (among others).
\end{itemize}

\subsection{Pipeline Representation Meta-Features}
To make the pipeline representation compact and extendable, we chose to represent the pipeline's topology as a sequence of words. Each type of primitive is represented using a unique fixed-length hash. Next, we generate a sequence of hashes to represent each pipeline, with the order of hashes determined by the pipeline's topology.

\begin{figure}[H]
    \centering
    \includegraphics[width=0.9\columnwidth]{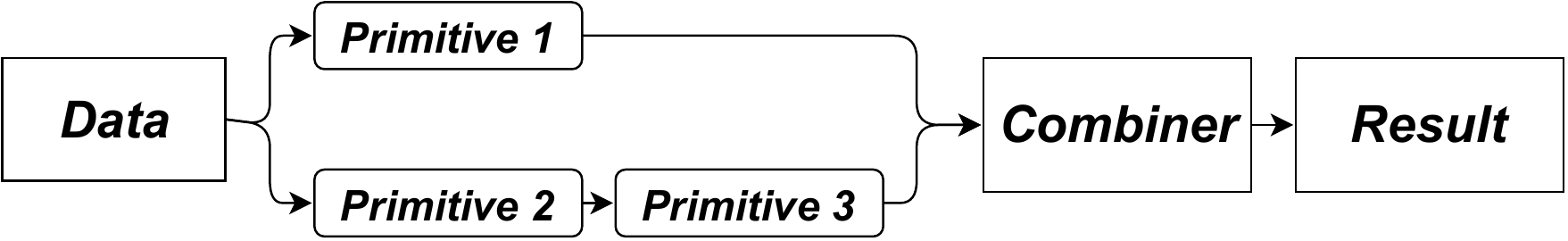}
    \caption[Pipeline simple representation]{An example of a pipeline.}
    \label{fig:pipeline_rep} 
\end{figure}

In order to make all pipeline representations consistent, we use the following rules to generate the representation:
\begin{itemize}
    \item We sequence the pipeline in reverse order -- from the final output to the inputs. A primitive must be sequenced prior to any of its input primitives. This is the case both for primitive with single inputs and multiple inputs (like the combiner primitive presented in Figure \ref{fig:pipeline_rep})
    \item In the case of multiple or parallel sub-pipelines (as in Figure \ref{fig:pipeline_rep}), the longest sub-pipeline is processed first. Ties are broken randomly.
    \item In order to make the representation equal in length for all pipelines, we define a fixed maximal number of primitives for all pipelines which is the number of primitives in the longest pipeline in the knowledge-base. In the case of smaller pipelines, padding (in the form of a designated ``blank'' primitive) is used.
\end{itemize}

\noindent An example of such transformation on a \textit{TPOT} based pipeline can be seen in figure \ref{fig:pipeline_rep}. Using our approach, the representation of the pipeline will be as follows:\\ \textit{[Combiner,Primitive3,Primitive2,data,Primitive1,data]}

\begin{figure*}[ht!]
    \centerline{\includegraphics[height=3.5in]{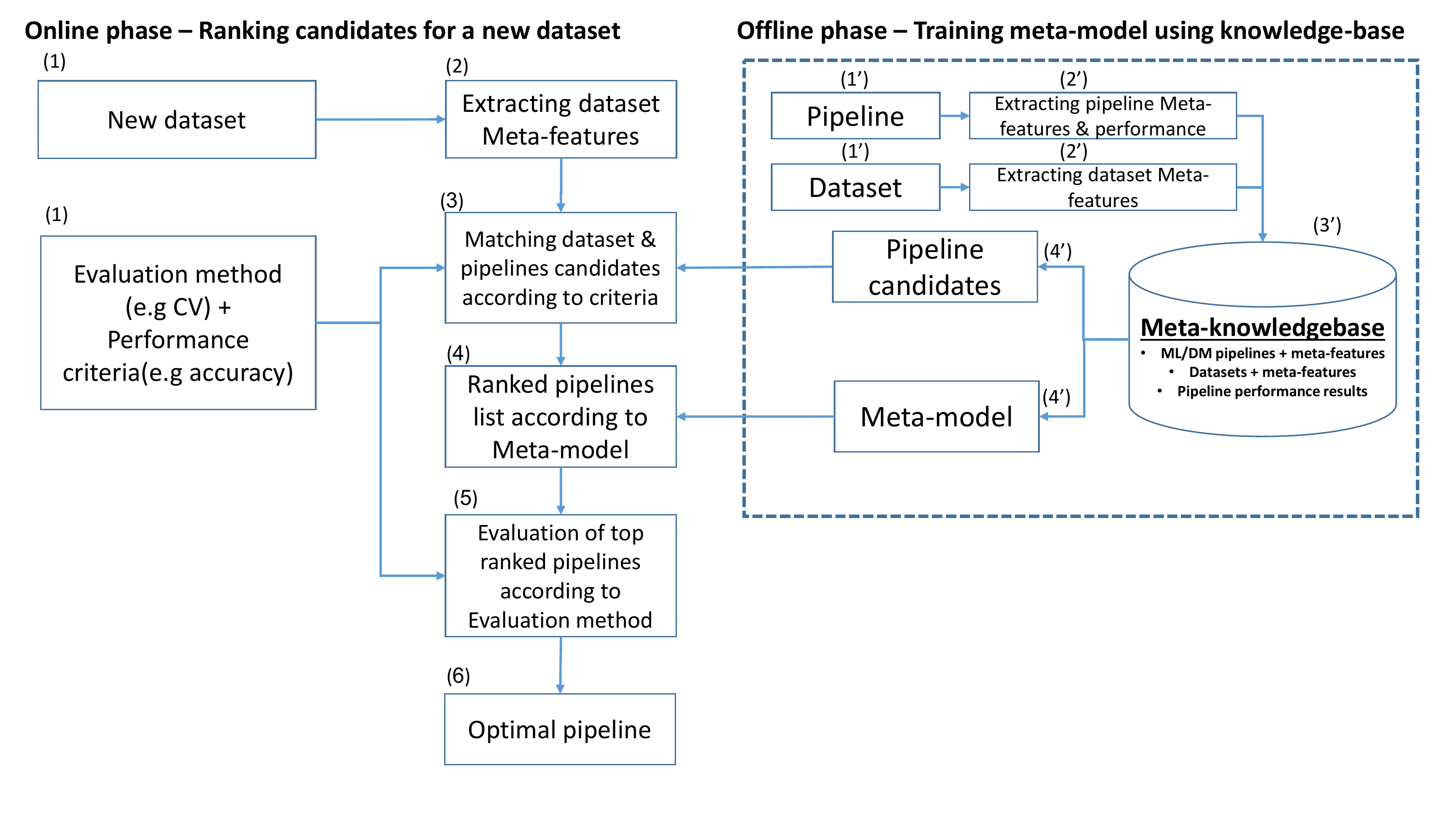}}
    \caption[Meta learner ranking]{The Meta-model work-flow. in the \textit{offline} phase, knowledge is gathered and a ranking algorithm is trained. in the \textit{online} phase, a new dataset and task are presented. meta features are extracted from the dataset and a list of candidate pipelines is acquire from the knowledge base according to the task at hand. the ranking model is used to ranked the pipelines and the optimal pipeline is recommended.}
    \label{fig:MetaLearning_workflow} 
\end{figure*}

\subsection{Training the Meta-Model}
Following the creation of the meta-features representing both the dataset and the various pipelines, we train the meta-model used for the pipeline ranking. The training of the meta-model is performed offline on a large knowledge base consisting of multiple datasets and ML pipeline architectures. The offline phase depicted in Figure \ref{fig:MetaLearning_workflow}, which provides an overview of the process.

The training process is carried out as follows: for each dataset $D$ in the knowledge base, we retrieve the set of all possible task $T$ (e.g., classification, regression) and evaluation metrics $M$ (e.g., AUC, accuracy). For each combination of $\{D,T,M\}$ where $t \in T$ and $m \in M$, we generate a large set of candidate pipelines $C$. We then train all combinations of $\{D,t,m,c\}$ where $c\in C$. The performance scores of all the pipelines (based on the relevant metric) are stored in the knowledge base for future use. Overall, the offline phase requires a week of running for all datasets (an average of 55 minutes per datasets).

Finally, for each task and evaluation metric, we train a ranking algorithm using the information gathered during the offline evaluation described above. For each evaluated dataset and pipeline combination, we extract their corresponding meta-features and concatenate them. The joined meta-features vectors are used to train the ranking algorithm. The goal of the algorithm is to produce a list of all participating pipelines, ordered by their respective performance on the dataset. Once the training of the meta model (a matter of several hours for all datasets) is done, online predictions (i.e., for new, previously-unseen datasets) are done almost immediately.
\section{Evaluation}
\subsection{Experimental Setup}
We evaluate our proposed approach on two common tasks in the field of machine learning: classification and regression. For each task, we assemble its own set of datasets and pipelines and train a separate meta-model (i.e., one for classification and one for regression).

\subsubsection{Datasets.} We used the 149 classification datasets and 95 regression datasets previously used in \cite{Noy2019}. These datasets are highly diverse with respect to their number of instances, number of features, feature composition, and class imbalance. All datasets are available in the following online repositories: UCI\footnote{https://archive.ics.uci.edu/ml},  
OpenML\footnote{https://www.openml.org} 
and Kaggle\footnote{www.kaggle.com}.

\subsubsection{Pipelines generation.} All the pipelines used in our training and evaluation were generated using TPOT \cite{Olson2019b},  a state-of-the-art framework for automatic pipeline generation and exploration. The pipelines generated by TPOT consist entirely of algorithms that can be found in the python \textit{scikit-learn} package. TPOT uses genetic algorithms to iteratively improve its generated pipelines. Moreover, TPOT supports the creation of parallel pipelines, an option that greatly increases the diversity of the pipelines population.

We ran TPOT on each of our datasets and collected all the architectures generated during runtime. We used TPOT's default settings -- $100$ pipelines per generation for $100$ generations with a default primitives dictionary consisting of $30$ primitives for classification tasks and $29$ primitives for regression tasks. This process resulted in an average of $9,700$ pipelines per dataset. Since TPOT sometimes generates the same pipelines for multiple datasets, we were able to obtain both pipelines that are unique to specific datasets and pipelines that are trained on multiple datasets. The former group provides our model with diversity, while the latter provides useful information on cross dataset-pipeline interactions.

While TPOT also performs hyper-parameter optimization in addition to its pipeline search, we consider this topic to be beyond the scope of our current work. Therefore, in cases where TPOT generated multiple pipelines with the same topology for a given dataset, we record only the performance of the top-performing pipeline. As a result, our knowledge base consisted of $142,006$ classification pipelines and $171,482$ regression pipelines. We make our entire database (datasets, pipeline architectures, and their performance) publicly available \footnote{the knowledge base and meta learner will be made available pending acceptance}.

It is important to note that while our current knowledge base is comprised solely from TPOT generated pipelines, all our meta-features are generic and can be applied to any type of ML pipeline representation. Our reasoning for using TPOT as the source of our pipelines is twofold: first, it is a state-of-the-art pipeline generation platform, so the chances of having at least some high-performing architectures to detect are high. Secondly, since we compare \MethodName's performance to those of TPOT and auto-sklearn, having our framework run only on pipelines generated from the same set of primitives ensures a fair comparison of all frameworks.

\subsubsection{Meta-learner implementation.} 
For the training of our meta-learner, we used XGBoost \cite{chen2016xgboost}. More specifically, we used the XGBRanker model with the pairwise ranking objective function and shallow trees of 150 estimators. This was done since our problem is in its essence a ranking problem, and previous work \cite{cao2007learning} has shown that XBGoost is highly suitable for producing ranked lists. Additionally, we used the following hyper-parameters settings: learning rate of 0.1, max depth of 8 and 150 estimators. We set the number of pipelines returned by \MethodName to $k=10$. The algorithm's parameters were empirically set using the leave-one-out approach. 

\subsubsection{Evaluation method.}
To test our ranking model we used a leave-one-out validation method. During the training phase, given the set of datasets $D$, we train a meta-model $\mathcal{M}_{d_{i}}$ using all datasets but ${d_{i}}$. This resulted in creating 149 different meta-models for classification and 95 for regression that were used in the experiment.

During the test phase, for each $d_i \in D$, we used the matching $\mathcal{M}_{d_{i}}$ meta-model to rank all possible pipelines and produce a ranked list based on predicted performance (see the online phase in Figure \ref{fig:MetaLearning_workflow}). The evaluated dataset $d_i$ was then split into train and test sets using a 80\%/20\% ratio. The $K$ top-ranked pipelines (by $\mathcal{M}_{d_{i}}$)  were then trained on the training set of $d_i$ and evaluated on its test set. The results of this evaluation were used to compare the performance of \MethodName and the baselines.

\subsubsection{Baselines.}
We compare the performance of \MethodName to those of the TPOT and auto-sklearn frameworks. Both frameworks are considered as state-of-the-art automatic pipeline generation platforms. For TPOT, we used the default settings, while for auto-sklearn we compared \MethodName to both a ``vanilla'' version which produces a single optimal pipeline (auto-sklearn(V)), and to a version which creates \textit{an ensemble} from the best 50 pipelines(auto-sklearn(E)). Obviously, the use of ensemble by auto-sklearn puts our approach at a significant disadvantage. In both cases, we limit auto-skelarn runtime to two hours and its memory limit to 8GB. It is important to stress again, however, that while both baseline frameworks evaluated each generated pipeline by running it on the evaluated dataset, \MethodName immediately produces a ranked list using its meta-model at a minimal computational cost.

\subsection{Results and Discussion}
\noindent \textbf{Classification results.}
Using the parameters of our experiments, the TPOT framework generated 10,000 pipelines for each dataset while auto-sklearn(v) and auto-sklearn(E) generated 3498 and 3520 pipelines on average respectively. All pipelines are then evaluated on the dataset's training set, and finally, a single pipeline or an ensemble is produced.  

\MethodName, on the other hand, utilizes the meta-model to rank \textit{all} the pipelines in the knowledge base with respect to the analyzed dataset and then returns its top-ranked pipelines. These pipelines are then trained on the datasets train set and evaluated on the test set. It is important to note that evaluating only the top-10 ranked pipelines makes \MethodName more efficient than TPOT by three orders of magnitude (10 to 10,000), and more efficient than auto-sklearn by two orders of magnitude. Since our ranking model is trained offline, the ranking is essentially without cost for new datasets (except for the meta-features extraction process, which takes only a few seconds).

The results of the evaluation on 149 datasets are presented in Tables \ref{table:1} and \ref{table:2}. It is clear that \MethodName's performances are comparable and even better than those of the baselines when $K\geq5$ even though it does not run any pipelines on the dataset prior to the recommendation.

\begin{table}[H]
\centering
\small
\begin{tabular}{|l|l|l|}
\hline
\textbf{Method}       & \textbf{Average Accuracy} & \textbf{stdev} \\ \hline
TPOT                  & 0.816                    & 0.159          \\ \hline
auto-sklearn(V)       & 0.796                      & 0.168          \\ \hline
auto-sklearn(E)        & 0.805                      & 0.165          \\ \hline
RankML \#1 rank       & 0.786                      & 0.169          \\ \hline
RankML Max top-5 rank &0.819                     & 0.154          \\ 
\hline
RankML Max top-10 rank &\textbf{0.827}                     & 0.152          \\ 
\hline
\end{tabular}
\caption[Average accuracy Results]{Average accuracy results across 149 classification datasets.}
\label{table:1}
\end{table}

\begin{table}[H]
\resizebox{\columnwidth}{!}{%
\begin{tabular}{|l|c|c|c|}
\hline
\multicolumn{1}{|c|}{\multirow{2}{*}{\textbf{Method}}} & \multicolumn{3}{c|}{\textbf{Number of Datasets with BOC Performance(\%)}} \\ \cline{2-4} 
\multicolumn{1}{|c|}{} & \textbf{TPOT} & \textbf{auto-sklearn(V)} & \textbf{auto-sklearn(E)} \\ \hline
RankML \#1 rank & 65(49\%) & 74(55\%) & 68(51\%) \\ \hline
RankML Max top-5 rank & 102(73\%) & 112(80\%) & 112(80\%) \\ \hline
RankML Max top-10 rank & \textbf{110(79\%)} & \textbf{119(85\%)} & \textbf{118(84\%)} \\ \hline
\end{tabular}%
}
\caption{The number of classification datasets each method got better or comparable(BOC) results against baselines (percentage is out of valid datasets).}
\label{table:2}
\end{table}

\begin{figure*}[ht]
	\centering
	\includegraphics[width=1.1\columnwidth]{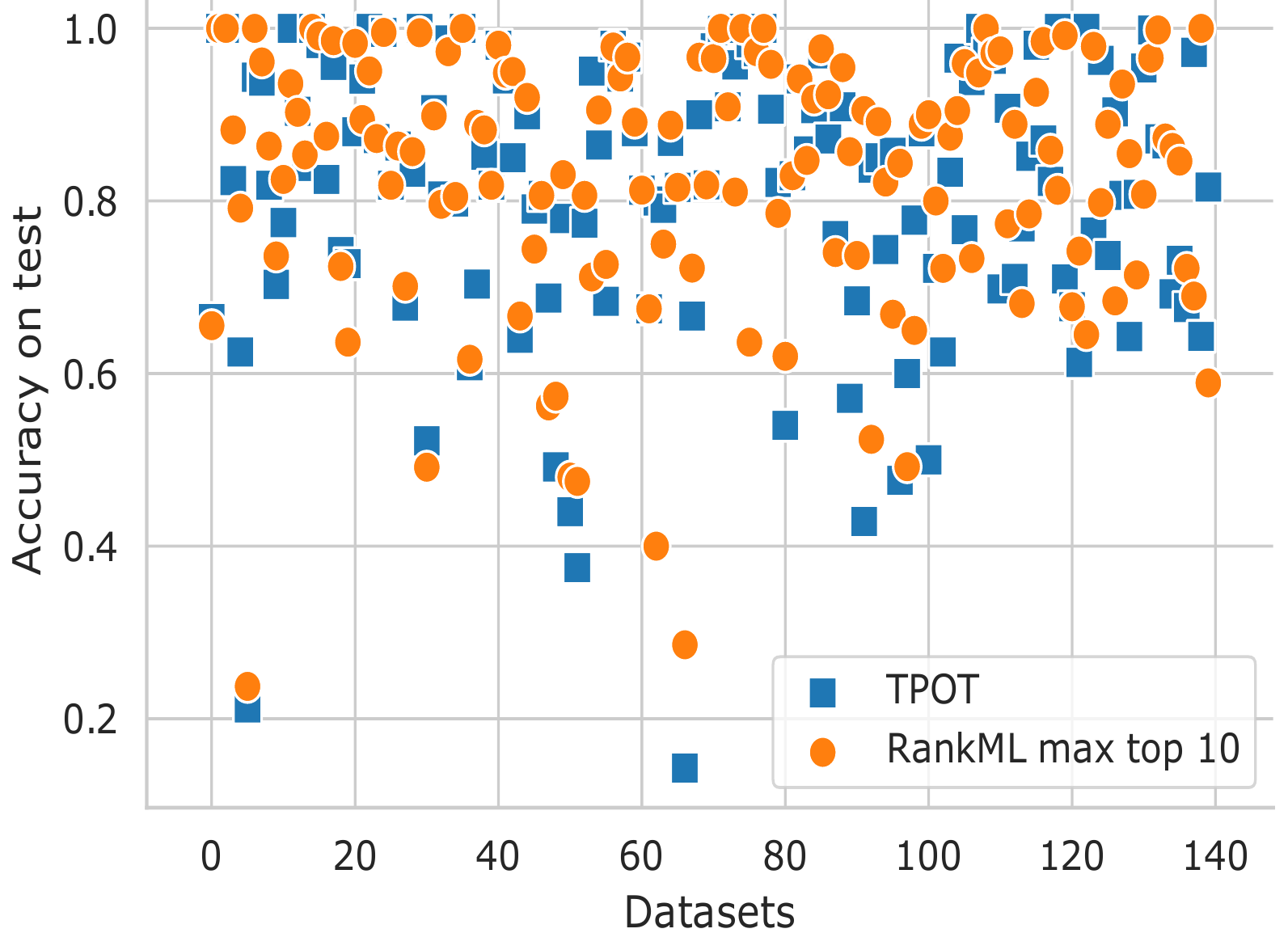}
	\caption[Accuracy compare]{A scatter diagram showing the accuracy on each of the 149 datasets. We present the results for RankML max of top-10 ranking as well as the "stronger" baseline TPOT.}
	\label{fig:scatter_acc_compare} 
\end{figure*}

\begin{figure}[H]
	\centering
	\includegraphics[width=0.9\columnwidth]{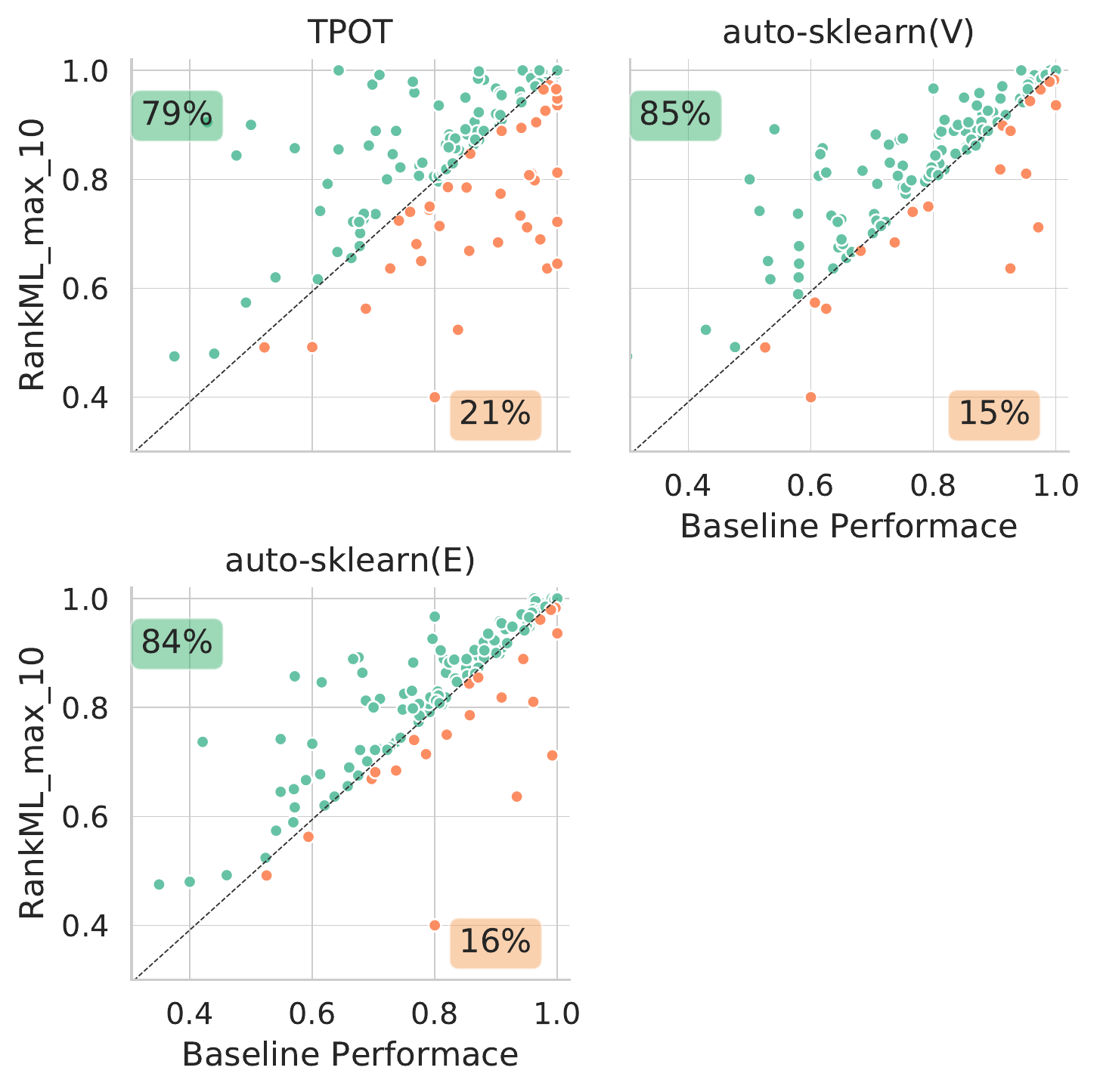}
	\caption[Box plot accuracy]{RankML best vs. Baselines  performances over  149 classification datasets. BOC percentage for each method is also indicated.}
	\label{fig:acc_boc_comapre} 
\end{figure}

\begin{figure}[t]
	\centering
	\includegraphics[width=0.9\columnwidth]{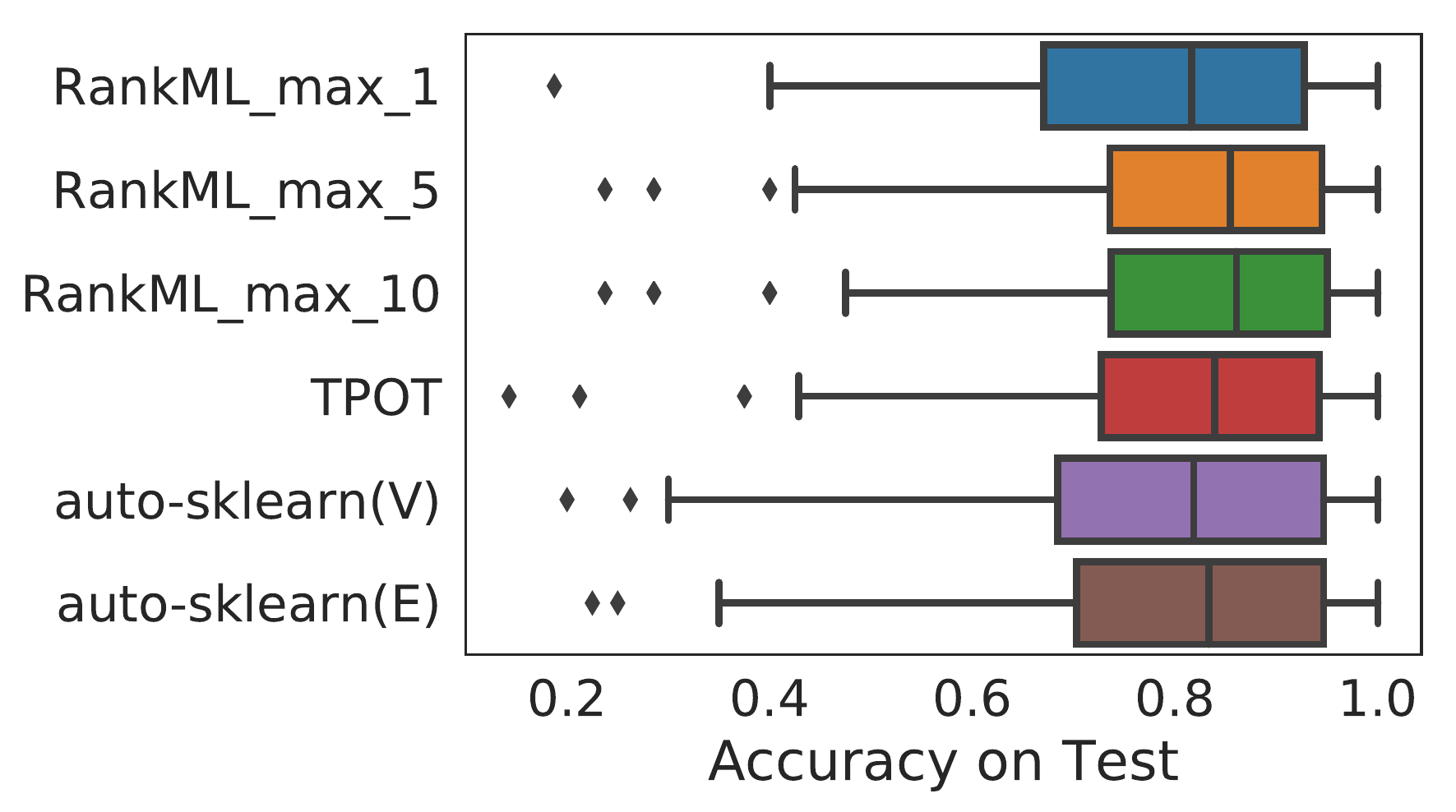}
	\caption[Box plot accuracy]{A boxplot showing the average test scores the baselines and RankML achieves when RankML is using the optimal pipeline out of k-top recommendations for 149 classification datasets.}
	\label{fig:boxplot_score} 
\end{figure}

\begin{figure}[ht]
	\centering
	\includegraphics[width=0.91\columnwidth]{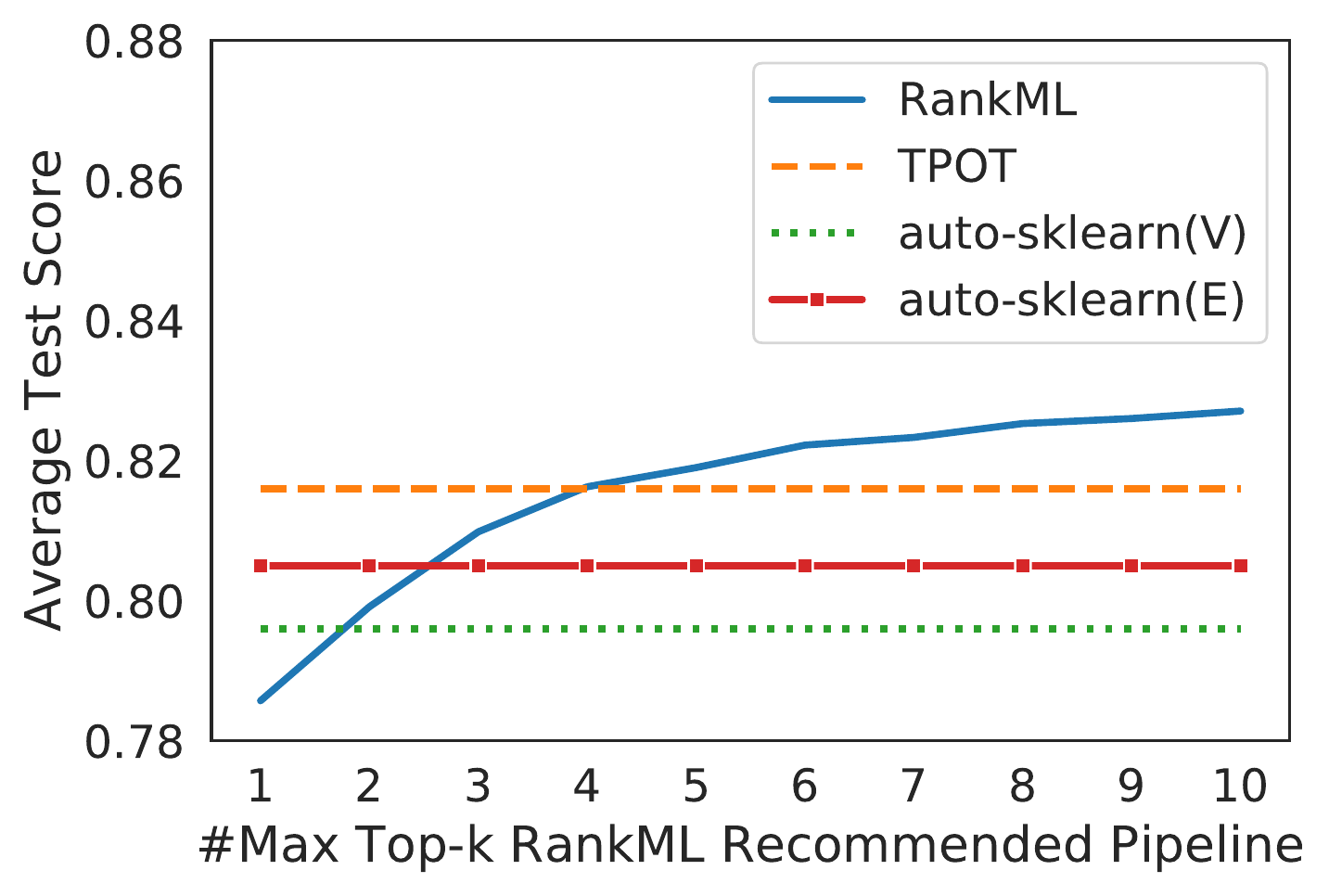}
	\caption[Avg RanKML max test]{A plot showing the average test scores RankML achieves when using the optimal pipeline out of k-top recommendations for 149 classification datasets.}
	\label{fig:line_avg_max_test} 
\end{figure}

\begin{figure*}[ht]
	\centering
	\includegraphics[width=1.97\columnwidth]{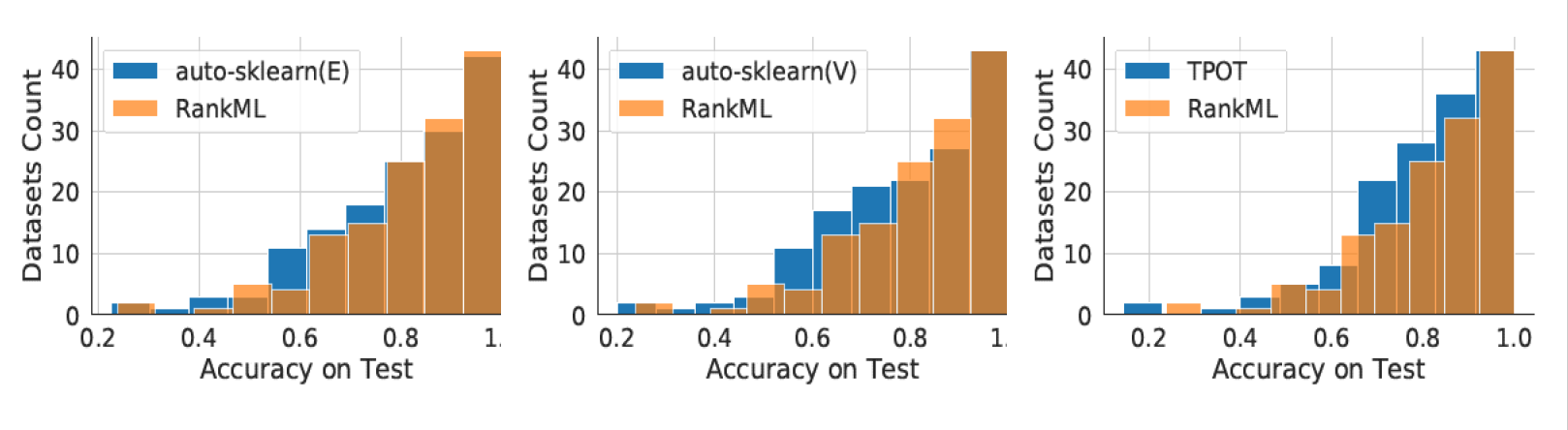}
	\caption[RankML hist]{Histograms showing accuracy scores across all classification datasets. We present the results for RankML against each of the baselines}
	\label{fig:hist_plot} 
\end{figure*}

Figures \ref{fig:scatter_acc_compare} - \ref{fig:hist_plot} provide further analysis of \MethodName's performance. Figure \ref{fig:scatter_acc_compare} presents the performance of  \MethodName and TPOT (the top-performing baseline) on all classification datasets while  Figure \ref{fig:acc_boc_comapre} plots the relative performance of our approach to each of the baselines.  It  is  clear  that  our  approach  outperforms the baselines in  a  large  majority of  cases,  with  the  percentage  of datasets in which there is an improvement  ranging  from  79\%  to  85\%.

 Figure \ref{fig:boxplot_score} presents boxplots of the accuracy score on the test set for each method, it clearly shows the merit of \MethodName when compared to the two state-of-the-art baselines.

Figure \ref{fig:line_avg_max_test} presents the average number of pipelines per dataset that \MethodName needs to evaluate in order to reach specific levels of performance (baselines performances are plotted as well). All analysis points to the fact that \MethodName achieves levels of performance that are very close to -- and sometimes surpass -- those of the baselines. Specifically, \MethodName needs to evaluate only four pipelines per dataset on average to reach a level of performance equal to that of the top-performing baseline---TPOT.

We use the Wilcoxon signed-rank test to determine whether the accuracy-based performance of our proposed approach outperforms that baselines in a statistically significant manner. Using a confidence level of 95\%, we were \textit{not} able to reject the null hypothesis for TPOT, but were able to reject it for both versions of auto-sklearn. This means that there is no significant difference in the performance of our approach to those of TPOT, but there are significant differences in the performance of our approach compared to those of auto-sklearn.

Finally, we analyzed the pipelines recommended by our approach. Our goal was to determine whether our recommendations were diverse (and therefore robust). More specifically, our goal was to verify that the meta-model didn't overfit the data by always recommending the same pipelines for any given dataset. Table \ref{table:4} presents the percentage of times the most frequent primitives were used in \MethodName's top-ranked pipelines. While it is clear that feature selection is a commonly used primitive type, the most common primitive (MaxAbsScaler) is only used in 16\% of the recommended pipelines. This leads us to conclude that \MethodName does indeed return a diverse set of pipelines.\newline

\begin{table}[H]
\centering
\resizebox{\columnwidth}{!}{%
\begin{tabular}{|m{3.3cm}|m{2cm}|m{1.1cm}|}
\hline
\multicolumn{1}{|c|}{\textbf{Primitive}} & \multicolumn{1}{c|}{\textbf{Family}} & \multicolumn{1}{c|}{\textbf{\begin{tabular}[c]{@{}c@{}}Avg \% \\ of appearances\end{tabular}}} \\ \hline
MaxAbsScaler & Data pre-processing & 16\% \\ \hline
StandardScaler & Data pre-processing & 11\% \\ \hline
KNeighborsClassifier & Predictive models & 10\% \\ \hline
RandomForestClassifier & Predictive models& 4\% \\ \hline
PCA & Feature pre-processing & 0.4\% \\ \hline
\end{tabular}%
}
\caption{A selection of primitives used in RankML recommended pipelines for classification. "Avg \% of appearances" is out of all primitives.
}
\label{table:4}
\end{table}

\noindent \textbf{Regression results.}
We conduct our evaluation of 95 datasets and use mean squared error (MSE) as our evaluation metric. The results of our evaluation are presented in Table \ref{table:1_reg}, which shows the percentage of dataset in which \MethodName achieved better-or-comparable performances to the baselines (depending on the number of evaluated pipelines). Its clear that \MethodName is able to reach comparable results against the baselines. Again, this result is particularly impressive given the fact that \MethodName does not conduct any evaluation on the analyzed dataset.

\begin{table}
\resizebox{\columnwidth}{!}{%
\begin{tabular}{|l|c|c|c|}
\hline
\multicolumn{1}{|c|}{\multirow{2}{*}{\textbf{Method}}} & \multicolumn{3}{c|}{\textbf{Number of Datasets with BOC Performance(\%)}} \\ \cline{2-4} 
\multicolumn{1}{|c|}{} & \textbf{TPOT} & \textbf{auto-sklearn(V)} & \textbf{auto-sklearn(E)} \\ \hline
RankML \#1 rank & 30(39\%) & 35(46\%) & 26(34\%) \\ \hline
RankML Max top-5 rank & 39(49\%) & 47(59\%) & 41(52\%) \\ \hline
RankML Max top-10 rank & \textbf{45(57\%)} & \textbf{53(67\%)} & \textbf{45(57\%)} \\ \hline
\end{tabular}%
}
\caption{The number of regression datasets each version of our approach got better or comparable results against the baselines 
(percentage is out of valid datasets).}
\label{table:1_reg}
\end{table}

\begin{figure}[h]
	\centering
	\includegraphics[width=0.99\columnwidth]{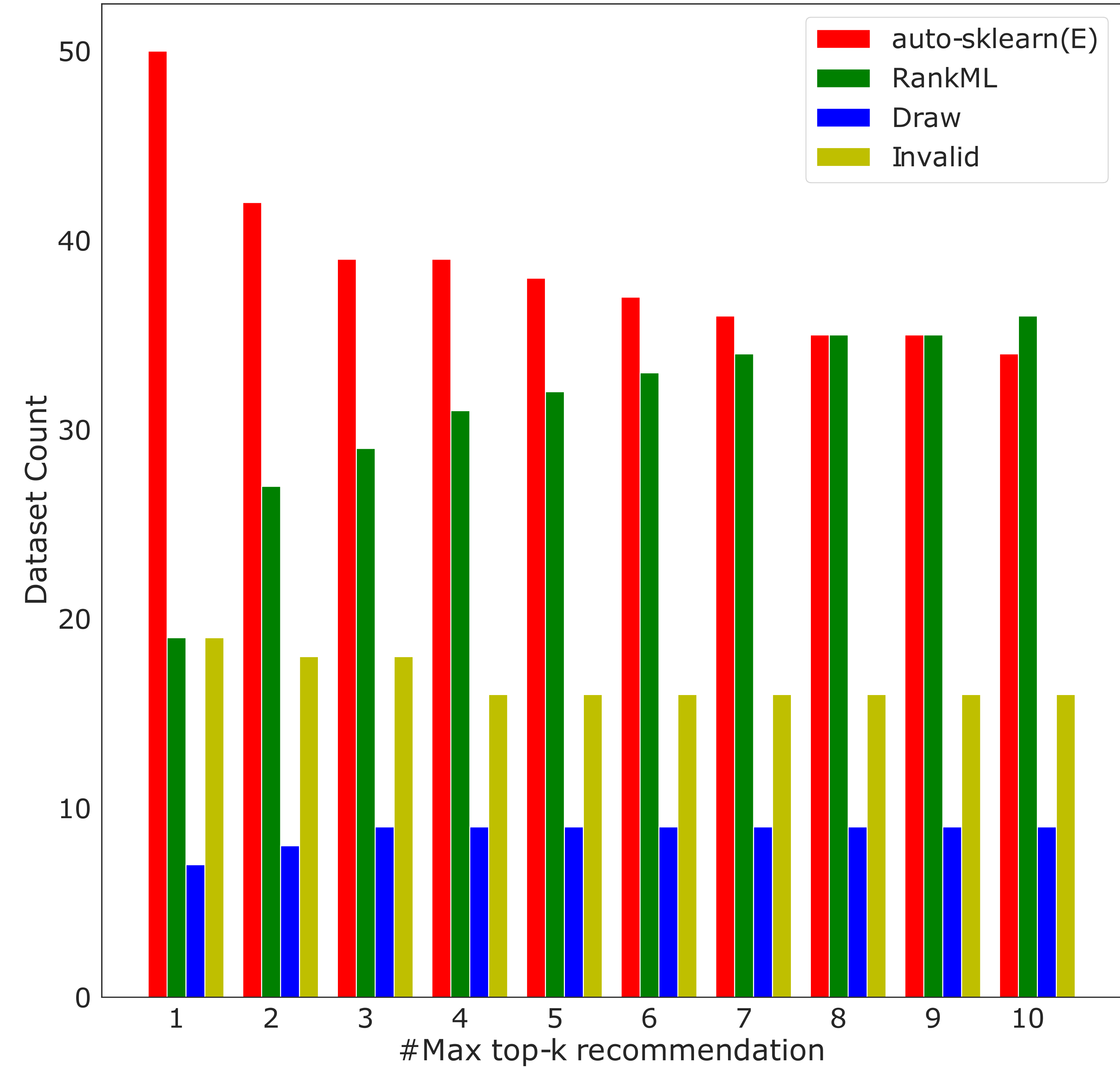}
	\caption[autosklaern vs max Meta test]{A bar plot showing the number of datasets for which each of the analyzed methods achieved top-performance as a function of $k$, the number of pipelines evaluated by \MethodName. For this analysis we use the top-ranked pipeline out of the $k$ that were evaluated. ``Draw'' represents cases where the differences among all methods were $\leq0.01$, while ``invalid'' represents cases where at least one baseline failed to complete its evaluation (mostly due to known issues in both TPOT and auto-sklearn). \footnote{https://github.com/EpistasisLab/tpot/issues/893}}
	\label{fig:bar_plot_max_test_reg} 
\end{figure}

Figure \ref{fig:bar_plot_max_test_reg} plots for each approach the number of dataset in which it outperformed the other as a function of $K$ (the number of top-ranked pipelines evaluated). The results show that \MethodName reaches comparable perforamnce to auto-sklearn(E) for $K\geq8$, which means we only have to evaluate eight pipelines on average to be \textit{comparable to an ensemble-based baseline}.

Finally, we used the Friedman test to determine \MethodName performances are indistinguishable from those of the baselines using a confidence level of 95\%.

\noindent \textbf{Discussion.} Our evaluation clearly shows that \MethodName is able to achieve results that are either comparable to the state-of-the-art (regression) or surpass it (classification). However, another significant metric is the required running time: while TPOT \textit{require 115 and 53} minutes per dataset on average for classification and regression respectively, \MethodName requires only \textit{three minutes and one minute}. auto-sklearn used its entire time limit of two hours to produce its results. These results are further proof to the effectiveness of our approach. 

While our approach is very effective it does require a large knowledge-base for inferring pipeline performance. As a result, our approach is less likely to perform well in cold-start use-cases and in scenarios where the analyzed dataset has characteristics that are significantly different from any previously seen dataset. It should be noted, however, that these shortcomings can be addressed by learning from a large and diverse set, as we have done in this study.

\section{Conclusion - Future work}
In this study, we presented RankML, a novel meta learning-based approach for ranking machine learning pipelines. By exploring the interactions between datasets and pipeline topology, we were able to train learning models capable of identifying effective pipelines without performing computationally-expensive analysis. By doing so, we address one of the main shortcomings of AutoML-based systems: long running times and computational complexity.

For future work, we plan to extend and test our method on different machine learning tasks. Additionally, we intend to explore more advanced meta-representations both for the datasets and pipelines. Finally, we intend to use our method as a step for improving existing AutoML systems.

\bibliography{ecai}

\begin{thebibliography}{10}

\bibitem{bergstra2012random}
James Bergstra and Yoshua Bengio, `Random search for hyper-parameter
  optimization', {\em Journal of Machine Learning Research}, {\bf 13}(Feb),
  281--305, (2012).

\bibitem{brazdil2008metalearning}
Pavel Brazdil, Christophe~Giraud Carrier, Carlos Soares, and Ricardo Vilalta,
  {\em Metalearning: Applications to data mining}, Springer Science \& Business
  Media, 2008.

\bibitem{brazdil2003ranking}
Pavel~B Brazdil, Carlos Soares, and Joaquim~Pinto Da~Costa, `Ranking learning
  algorithms: Using ibl and meta-learning on accuracy and time results', {\em
  Machine Learning}, {\bf 50}(3),  251--277, (2003).

\bibitem{cao2007learning}
Zhe Cao, Tao Qin, Tie-Yan Liu, Ming-Feng Tsai, and Hang Li, `Learning to rank:
  from pairwise approach to listwise approach', in {\em Proceedings of the 24th
  international conference on Machine learning}, pp. 129--136. ACM, (2007).

\bibitem{chandola2009anomaly}
Varun Chandola, Arindam Banerjee, and Vipin Kumar, `Anomaly detection: A
  survey', {\em ACM computing surveys (CSUR)}, {\bf 41}(3), ~15, (2009).

\bibitem{chen2018autostacker}
Boyuan Chen, Harvey Wu, Warren Mo, Ishanu Chattopadhyay, and Hod Lipson,
  `Autostacker: A compositional evolutionary learning system', {\em arXiv
  preprint arXiv:1803.00684}, (2018).

\bibitem{chen2016xgboost}
Tianqi Chen and Carlos Guestrin, `Xgboost: A scalable tree boosting system', in
  {\em Proceedings of the 22nd acm sigkdd international conference on knowledge
  discovery and data mining}, pp. 785--794. ACM, (2016).

\bibitem{Noy2019}
Roman Vainshtein Gilad Katz Rokach~Lior Cohen-Shapira, Noy, `Autogrd: Model
  recommendation through graphical dataset representation', in {\em Proceedings
  of the 28th international conference on Machine learning}. ACM, (2019).

\bibitem{covington2016deep}
Paul Covington, Jay Adams, and Emre Sargin, `Deep neural networks for youtube
  recommendations', in {\em Proceedings of the 10th ACM conference on
  recommender systems}, pp. 191--198. ACM, (2016).

\bibitem{davenport2012data}
Thomas~H Davenport and DJ~Patil, `Data scientist', {\em Harvard business
  review}, {\bf 90}(5),  70--76, (2012).

\bibitem{Drori2018b}
Iddo Drori, Yamuna Krishnamurthy, Remi Rampin, Raoni De, Paula Lourenco,
  Jorge~Piazentin Ono, Kyunghyun Cho, Claudio Silva, and Juliana Freire,
  `{AlphaD3M: Machine Learning Pipeline Synthesis}', {\em JMLR Work. Conf.
  Proc.}, {\bf 1},  1--8, (2018).

\bibitem{feurer2018practical}
Matthias Feurer, Katharina Eggensperger, Stefan Falkner, Marius Lindauer, and
  Frank Hutter, `Practical automated machine learning for the automl challenge
  2018', in {\em International Workshop on Automatic Machine Learning at ICML},
  pp. 1189--1232, (2018).

\bibitem{feurer2018scalable}
Matthias Feurer, Benjamin Letham, and Eytan Bakshy, `Scalable meta-learning for
  bayesian optimization using ranking-weighted gaussian process ensembles', in
  {\em AutoML Workshop at ICML}, (2018).

\bibitem{Feurer2015d}
Matthias Feurer, Jost~Tobias Springenberg, and Frank Hutter, `{Initializing
  Bayesian Hyperparameter Optimization via Meta-Learning}', {\em Aaai},
  1128--1135, (2015).

\bibitem{Feurer2015e}
Matthias Feurer, Jost~Tobias Springenberg, Aaron Klein, Manuel Blum, Katharina
  Eggensperger, and Frank Hutter, `{Efficient and Robust Automated Machine
  Learning}', {\em Proc. 28th Int. Conf. Neural Inf. Process. Syst.},
  2755--2763, (2015).

\bibitem{fusi2018probabilistic}
Nicolo Fusi, Rishit Sheth, and Melih Elibol, `Probabilistic matrix
  factorization for automated machine learning', in {\em Advances in Neural
  Information Processing Systems}, pp. 3348--3357, (2018).

\bibitem{jin2019auto}
Haifeng Jin, Qingquan Song, and Xia Hu, `Auto-keras: An efficient neural
  architecture search system', in {\em Proceedings of the 25th ACM SIGKDD
  International Conference on Knowledge Discovery \& Data Mining}, pp.
  1946--1956. ACM, (2019).

\bibitem{Katz2017b}
Gilad Katz, Eui Chul~Richard Shin, and Dawn Song, `{ExploreKit: Automatic
  feature generation and selection}', {\em Proc. - IEEE Int. Conf. Data Mining,
  ICDM},  979--984, (2017).

\bibitem{khalid2014survey}
Samina Khalid, Tehmina Khalil, and Shamila Nasreen, `A survey of feature
  selection and feature extraction techniques in machine learning', in {\em
  2014 Science and Information Conference}, pp. 372--378. IEEE, (2014).

\bibitem{Lemke2015b}
Christiane Lemke, Marcin Budka, and Bogdan Gabrys, `{Metalearning: a survey of
  trends and technologies}', {\em Artif. Intell. Rev.}, {\bf 44}(1),  117--130,
  (2015).

\bibitem{Luo2016b}
Gang Luo, `{A review of automatic selection methods for machine learning
  algorithms and hyper-parameter values}', {\em Netw. Model. Anal. Heal.
  Informatics Bioinforma.}, {\bf 5}(1),  1--16, (2016).

\bibitem{milutinovic2017end}
Mitar Milutinovic, At{\i}l{\i}m~G{\"u}ne{\c{s}} Baydin, Robert Zinkov, William
  Harvey, Dawn Song, Frank Wood, and Wade Shen, `End-to-end training of
  differentiable pipelines across machine learning frameworks', (2017).

\bibitem{Olson2019b}
Randal~S. Olson and Jason~H. Moore, `{TPOT: A Tree-Based Pipeline Optimization
  Tool for Automating Machine Learning}',  151--160, (2019).

\bibitem{Pfahringer2000b}
Bernhard Pfahringer, Hilan Bensusan, and Christophe Giraud-Carrier,
  `{Meta-Learning by Landmarking Various Learning Algorithms}', {\em Proc.
  Seventeenth Int. Conf. Mach. Learn. ICML2000}, {\bf 951}(2000),  743--750,
  (2000).

\bibitem{10.1007/978-3-319-31753-3_18}
F.~Pinto, C.~Soares, and J.~Mendes-Moreira, `Towards automatic generation of
  metafeatures', in {\em Pacific-Asia}, pp. 215--226, (2016).

\bibitem{doi:10.1002/widm.1249}
Omer Sagi and Lior Rokach, `Ensemble learning: A survey', {\em Wiley
  Interdisciplinary Reviews: Data Mining and Knowledge Discovery}, {\bf 8}(4),
  e1249, (2018).

\bibitem{Santoro2016b}
Adam Santoro, Sergey Bartunov, Matthew Botvinick, Daan Wierstra, Timothy
  Lillicrap, and Google Deepmind, `{Meta-Learning with Memory-Augmented Neural
  Networks Google DeepMind}', {\em Jmlr}, {\bf 48},  1842--1850, (2016).

\bibitem{Thornton2012b}
Chris Thornton, Frank Hutter, Holger~H. Hoos, and Kevin Leyton-Brown,
  `{Auto-WEKA: Combined Selection and Hyperparameter Optimization of
  Classification Algorithms}', (2012).

\bibitem{Vainshtein2018}
Roman Vainshtein, Asnat Greenstein-Messica, Gilad Katz, Bracha Shapira, and
  Lior Rokach, `{A Hybrid Approach for Automatic Model Recommendation}',
  1623--1626, (2018).

\bibitem{Vilalta2002b}
Ricardo Vilalta and Youssef Drissi, `{A perspective view and survey of
  meta-learning}', {\em Artif. Intell. Rev.}, {\bf 18}(2),  77--95, (2002).

\bibitem{wang2015link}
Peng Wang, BaoWen Xu, YuRong Wu, and XiaoYu Zhou, `Link prediction in social
  networks: the state-of-the-art', {\em Science China Information Sciences},
  {\bf 58}(1),  1--38, (2015).

\bibitem{wolpert2002supervised}
David~H Wolpert, `The supervised learning no-free-lunch theorems', in {\em Soft
  computing and industry},  25--42, Springer, (2002).

\end{thebibliography}
\end{document}